# IMPLEMENTATION AND COMPARATIVE QUANTITATIVE ASSESSMENT OF DIFFERENT MULTISPECTRAL IMAGE PANSHARPENING APPROACHES

Shailesh Panchal[1] and Dr. Rajesh Thakker[2]

[1]Phd Scholar, Department of Computer Engineering, CHARUSAT, Changa, Gujarat
[2]Professor and Head, Department of Electronics Communication,
VGEC Chandkheda, Ahmadabad, Gujarat

## ABSTRACT

*In remote sensing, images acquired by various earth observation satellites tend to have either a high spatial and low spectral resolution or vice versa. Pansharpening is a technique which aims to improve spatial resolution of multispectral image. The challenges involve in the pansharpening are not only to improve the spatial resolution but also to preserve spectral quality of the multispectral image. In this paper, various pansharpening algorithms are discussed and classified based on approaches they have adopted. Using MATLAB image processing toolbox, several state-of-art pan-sharpening algorithms are implemented. Quality of pansharpened images are assessed visually and quantitatively. Correlation coefficient (CC), Root mean square error (RMSE), Relative average spectral error (RASE) and Universal quality index (Q) indices are used to measure spectral quality while to spatial-CC (SCC) quantitative parameter is used for spatial quality measurement. Finally, the paper is concluded with useful remarks.*

## 1. INTRODUCTION

Nowadays, various earth observation satellites such as IKONOS, Quickbird, SPOT, Landsat, etc. provide images at different spatial, temporal and spectral resolutions [1]. The spatial resolution of image is expressed as area of the ground covered by one pixel of the image. As pixel size is reduced, objects in the image are delineated with high accuracy. The instantaneous field of view (IFOV) is the portion of the ground which is sensed by the sensor. Spatial resolution depends on the IFOV. As finer the IFVO, spatial resolution is better, and objects in the image can be classified with more accuracy [2]. For example, the LANDSAT-7 satellite has capability to capture the image with 15m spatial resolution while GeoEye-I satellite provides 0.41m spatial resolution. Normally, less than 4m pixel size is considered as high spatial resolution while pixel size of more than 30m, is considered as low spatial resolution. Spectral resolution is characterized by reflectance over a variety of signal wavelength [3]. Spectral resolution is higher if bandwidth is narrower [3]. A Panchromatic (PAN) image contains one band of reflectance data that covers a broad spectral range and while maintaining a high signal-to-noise ratio, this allows smaller detectors to be utilized. Therefore, a PAN image has usually low spectral resolution, but high spatial resolution [4]. The principal category of images obtained by remote sensing is the Multi Spectral images. An MS image contains more than one band and for the most part this band uses three-band. The spectral range of every band of a MS image is not as much as that of the PAN image, resulting about high spectral resolution, but low spatial resolution [5]. Numerous remote sensing satellites acquire images in one PAN band of high spatial resolution and a few MS bands of high spectral resolution. Image fusion technology used effectively in a wide variety of this field and has turned into an effective solution for expanding prerequisites for images of high spatial and high spectral resolution at the same time and this method otherwise called as pansharpening. However, Pansharpening is a procedure of merging high-resolution panchromatic and lower resolution Multi Spectral images to make a single high-resolution color image. It alludes to a sharpening procedure using the PAN band and a procedure of merging high-resolution panchromatic. A variety of image fusion systems dedicated to combining multi spectral and panchromatic images [6], [7], [8], [9]. Image fusion is the procedure of joining high spatial resolution panchromatic (PAN) image and rich Multi Spectral (MS) image into a single image. Motivation behind pansharpening is to obtaining information of greater quality and a vital tool for information enhancement, spatial resolution improvement, multi-data integration, and change detection. In recent years, numerous image fusion systems, such as, principal component analysis, intensity-hue-saturation, Brovey transforms and multi-scale transforms, etc., have been proposed to intertwine the PAN and MS images successfully. Most of the consideration paid to image enhancement with distinctive remote sensing images. With such images, particularly for image interpretation or classification, it would be vastly improved to utilize all the information contained in the original data, instead of getting an optimum image display with other extravagant high spatial resolution images. However, there is still no pertinent technique to enhance the spatial information in these images, without losing their spectral resolution.

This paper is organized as follows. Pre-processing is discussed in Section 2 whereas different state-of-the-art pansharpening techniques are described in Section 3. Subjective and objective quality assessment parameters used to measure the spectral and spatial quality of the resultant pansharpened image are presented in Section 4. Results are provided in Section 5. Finally, conclusions are drawn in Section 6.

## 2. PREPROCESSING

PAN and MS images are to be pre-processed before pansharpening. Pre-processing may involve image registration, resampling and histogram matching of the input images. Pre-processing techniques before pansharpening is broad area of research [10]. Initially, in image registration, input images are adjusted for spatial alignments such that pixels in the input images refer to the same points and objects on the ground. It is followed by resizing of multispectral images to that of panchromatic image using the interpolation or by different up-sampling techniques [11]. In some cases, histogram matching is performed before applying pansharpening techniques. Histogram matching of MS and PAN images may reduce spectral distortion in the resultant pansharpened image [10, 11].

# 3. PANSHARPENING STATE-OF-ART TECHNIQUES

Pansharpening is performed at three different processing levels [12]. These are pixel, feature and decision levels. Image pansharpening techniques are classified in several ways. In [13], classifications are provided on the basis of spatial domain, spectral domain and scale-space techniques. In [14], pansharpening techniques are classified in different families such as component substitution (CS), relative spectral contribution and high-frequency injection. Pansharpening techniques can be categorized based on the approaches used by them. The following are the some identified categories.

## 3.1. Component Substitute technique

Multispectral (MS) images generally contain more than one band, i.e., red, green and blue visible bands. In the component substitute category, MS image is transformed into a set of components using linear transform techniques. PAN image contains single band, and it contains a high-frequency component. Before, low-frequency component of the MS image is substituted with a high-frequency component of the PAN image, MS image is required to be resized to that of PAN image. Generally this substitution occurs in the transformed domain. In [15, 16], several component substitute approaches are discussed. In this study, it is assumed that MS and PAN images are registered images. The steps followed in this approach are as follows:

1. Carry out up-sampling to increase the size of MS image to that of PAN image.

2. Perform the forward transform to separate spatial and spectral components of the MS image.

3. Substitute the spatial components of MS image with that of the histogram matched PAN image.

4. Perform the backward transform to obtain the MS image back with the improved spatial resolution called pansharpened image.

The various pansharpening algorithms like Intensity-Hue-Saturation (IHS) [17], adaptive IHS [18], and Principal Component Analysis (PCA) [19] are the examples of the components substitute family.

In Intensity-Hue-Saturation, IHS colour space is used because it very well separates the intensity component (I) and spectral components (H and S) from the input MS images. Intensity (I) represents the total luminance of the image, hue represents the dominant wavelength contributing to the colour, and saturation describes the purity of the colour relative to grey. The basic idea of IHS transform is to replace the intensity component (I) of MS image by that of the histogram matched PAN image. The RGB of the resultant merged MS image is obtained by computing reverse IHS to RGB transform. The intensity band I calculated using following equation.

$$I = \sum_{i=1}^{N} \alpha_i * M_i$$

Here, $M_i$ are the multispectral image bands, and N is the number of bands. The value of coefficient $\alpha_i$ is taken 1/3 for N=3. Image captured with more than three bands, like IKONOS images, the value of α is experimentally required to be determined. To calculate the adaptive value of α, based on the number of bands available for the MS image, an approach known as adaptive IHS is proposed in [18]. In adaptive IHS approach, a value of α is determined such that the intensity (I) band approximates the corresponding PAN image as closely as possible. The mathematical formulation to determine adaptive values of α is as given below.

$$ I = \sum_{i=1}^{N} \alpha_i * M_i \approx PAN $$

Another method, principal component analysis (PCA) also falls into component substitute pansharpening category. PCA is basically mathematical model transformation [19-21]. It is widely used in the statistical application as well as signal processing area. In PCA, multivariate data sets with the correlated variables are transformed into a data set with new uncorrelated variables. Mostly, 1st principal component contains highest variance and it contains the maximum amount of information from the original image. [20].

## 3.2. Spectral contribution pansharpening technique

Another pansharpening technique, named Brovey transform (BT) is discussed in [22-23]. In BT, all the spectral bands are contributing equally to get the pansharpened image. It is based on the chromaticity transform. It is basically injecting the overall brightness of PAN image into each pixel values of normalized MS bands. Computations for each band are carried out as follow:

$$ R_{new} = \frac{R}{R + G + B} * PAN $$

$$ G_{new} = \frac{G}{R + G + B} * PAN $$

$$ B_{new} = \frac{B}{R + G + B} * PAN $$

Brovey transform provides good contrast visibility, but it greatly distorts the spectral characteristics [24]. It gives satisfactory performance when MS image contains only three bands.

## 3.3. High-frequency injection technique

The main idea in the case of high-frequency injection technique is to extract the high-frequency information from the PAN image by using high-pass filtering (HPF), and later inject it into the MS image. General algorithm is as follow:

1. Do the up-sampling process to increase the size of MS image to that of PAN image.

2. Apply the high-pass filter to PAN image.

3. Obtain the pansharpened image by adding resultant high-pass filtered PAN image to each of the bands of MS image.

The mathematical model for this family is as below:

$$MS_{high} = MS_{low} + PAN_{high}$$

This method preserves a great amount of spectral characteristics of the MS image since spectral information is associated with the low spatial frequency of the MS image. The cut-off frequency of the high-pass filtering is influence the spectral information of the MS image. Some recently reported methods are uses this high-frequency method as a predecessor to extract the spatial detail from the PAN image which is not present in the MS image.

### 3.4. Statistical technique

The pansharpening techniques based on the statistics explores statistical characteristics of the MS and PAN images. In [25], the price proposed a first statistical based approach called price method for pansharpening. Later, it was improved by Park et al. [26] with a spatially adaptive algorithm. In the price method, all high resolution (HR) pixels are modelled as a linear weight by some factor to one low-resolution (LR) pixel and due to this assumption; sometimes it is producing blocking artifacts effect. Spatially adaptive algorithm [26] was proposed to overcome this limitation and it used the adaptive approach to finding out the local correlation between the pixels resolutions in the input images. Besides these, Bayesian method was proposed based on the probability theory for estimation the final pansharpened image [27]. In Frosti et al. [28], pansharpening is considered as an ill-posed problem that needs regularization for optimal results. Hence, they chose total variation (TV) regularization model which produces the pansharpened image with preserving the fine details of PAN image.

### 3.5. Multi-resolution analysis technique

Multi-resolution analysis (MRA) based approach basically allows the spatial transformation over the wide range of scale instead of local processing of the pixels. Wavelet transforms [29], contourlet transform [30], curvelets transform and laplacian pyramid transform [31] are the multi-resolution based image pansharpening techniques. These techniques decomposed the MS and PAN images into different scale levels in order to derive the spatial information and importing it into finer scales of the MS images.

Wavelet provides a framework for the decomposition of the images with hierarchical degraded resolution and separating the spatial resolution detail at each level. The special case of MRA is discrete wavelet transform (DWT). Pansharpening approaches discussed in [32] proposed by Mallat's and "a' trous" are the examples of DWT. Mallat's approach for pansharpening is an orthogonal, non-symmetric, decimated and non-redundant while "a' trous" approach is non-orthogonal, symmetric, undecimated and redundant. Contourlet transform (CT) is proposed in [33]. DWT has very poor directional sensitivity and usually it is providing four subimages, which are referred to as LL, LH, HL and HH images. Contourlet transform (CT) is an alternative multi-resolution method which provides an efficient directional representation and capturing the intrinsic geometrical structures of the natural image along with smooth contours. In CT, transformation stage includes two filter banks: the Laplacian pyramid to generate multi-scale decomposition and the directional filter bank (DFB) to reveal directional details at each decomposition level. These methods, generally, following the steps as listed below:

1. Apply the forward transform to PAN and MS images using a sub-band and directional decomposition wavelet/contourlet transform.

2. Apply the fusion rule onto transform coefficients.

3. Obtain the pansharpened image by the inverse transform.

Fusion rules in the step 2, involves the substitution of original MS coefficient bands by coefficients of the PAN images or addition of these coefficients with some weight depending upon the contribution of the PAN and MS bands. Based on the applied fusion rules, they are also known as additive wavelet/contourlet or substitute wavelet/contourlet methods. Sometimes a hybrid approach is also used by combining best aspects of various fusion rules.

### 3.6. Other techniques

In spite of number of approaches to achieve the good quality of pansharpened image as discussed earlier, H.Yin et al. [34], proposed a novel framework for simultaneous image fusion and super-resolution. It is based on the sparse representation of the signal. It consists of three steps. First, low-resolution input images are interpolated and decomposed into high and low-frequency components. Second sparse coefficients are computed and finally fused image is achieved using fusion rules. In [35], Y. Zhang presented non-RIP based analysis technique for compressive sensing using $\ell_1$ −minimization. Compressive sensing (CS) theory has recently attracted intensive research activities in various fields. Conventional data acquisition can be called "full sensing and then compression" while compressive sensing means to reduce the number of measurements during data acquisition so that no additional compression is required. Finally, data is recovered from the measurement at the receiver side. Non-RIP based analysis technique improves the recoverability and stability compare to RIP based analysis.

## 4. QUALITY ASSESSMENTS

It is desirable to improve the spatial resolution of the MS image to that of PAN. Wald et al. [36] formulated some useful properties to verify the quality of the pansharpened image. They are

1. If pansharpened image is downsampled to its original spatial resolution then it should be similar to original MS image.

2. Pansharpened image should be as similar as possible to MS image which could be captured by the sensor (assuming that it is available) having high spatial resolution capacity.

The first property represents consistency property while, the second represents synthesis property. Quality of pansharpened image is measured against ideal reference image if reference image is available. Otherwise quality can be measured against input MS image called non-reference based quality assessment. Normally, the latter approach is followed.

### 4.1. Visual assessment

Generally image quality measured through visual inspections, are the global quality of the image like geometric shape, the size of the objects, spatial detail and local contrast of the image. By comparing the pansharpened image with input MS image, it is possible to verify and observe the spectral (color information) and spatial (sharpness) quality of the image. Visual assessment technique is a subjective technique.

## 4.2. Quantitative assessment

A number of parameters used for quantitative assessment of pansharpened image. In this paper, quantitative parameter values of pansharpened image are calculated against the input MS image. Here, R and F representing MS image and resultant pansharpened images, respectively.

### 4.2.1. Spectral quality assessment

To measure the spectral quality of pansharpened image, following quantitative parameters can be used [12]

1) *Correlation coefficient (CC)*: It indicates the spectral integrity of the pansharpened image [32]. It is calculated globally for the entire image. CC between pansharpened image F and input MS image R is computed as under:

$$CC(R, F) = \frac{\sum_{mn} (R_{mn} - \overline{R})(F_{mn} - \overline{F})}{\sqrt{(\sum_{mn}(R_{mn} - \overline{R})^2) (\sum_{mn}(F_{mn} - \overline{F})^2) )}}$$

Here, $\overline{F}$ and $\overline{R}$ are the mean value of the images F and R, respectively, while m and n is size of images. Value of CC should be as close to 1 as possible.

2) *Root mean square error (RMSE)*: It measures the changes in the radiance of the pixel values for each band of the input MS image R and pansharpened image F. It is important indicator when images under consideration contain homogeneous regions. It should be as close to zero as possible. RMSE is computed as follows:

$$RMSE = \sqrt{\frac{1}{m \times n} \sum_{i=1}^{m} \sum_{j=1}^{n} |R(i,j) - F(i,j)|^2}$$

3) *Relative average spectral error (RASE)*: It is computed using the root mean square error (RMSE) as per the below given equation.

$$RASE = \frac{100}{\mu} \sqrt{\frac{1}{N} \sum_{i=1}^{N} RMSE^2(B_i)}$$

Here, $\mu$ is the mean radiance of the N spectral bands and $B_i$ represents ith band of input MS image. The desired value of this parameter is zero.

4) *Universal quality index (UQI)*: In [37], image quality index suggested for the final pansharpened image F with respect to the input MS image R is as given below:

$$Q(R, F) \triangleq \frac{4\sigma_{RF} * \overline{R} * \overline{F}}{(\sigma_R^2 + \sigma_F^2)(\overline{R^2} + \overline{F^2})}$$

The above equation can also be rewritten as

$$Q(R, F) \triangleq \frac{\sigma_{RF}}{\sigma_{R*}\sigma_F} \times \frac{2 * \overline{R} * \overline{F}}{(\overline{R^2} + \overline{F^2})} \times \frac{2 * \sigma_{R*}\sigma_F}{(\sigma_R^2 + \sigma_F^2)}$$

The first factor indicates the correlation coefficient (CC) between images R and F, while the second factor indicates the luminance distance, and the third factor represents the contrast distortion between two images. In above equation, $\sigma_{RF}$ denotes the covariance between images R and F, $\overline{R}$ and $\overline{F}$ are the means while, $\sigma_R^2$ and $\sigma_F^2$ signify the standard deviation of R and F, respectively. The best value of Q is 1 and it can be achieved if R = F for all pixels.

5) Relative dimensionless global error in synthesis (ERGAS): The ERGAS is a global quality index and sensitive to mean shifting and dynamic range change. The value of the ERGAS indicates the amount of spectral distortion in the image.

$$ERGAS = 100 * \frac{h}{l} \sqrt{\frac{1}{N} \sum_{i=1}^{N} \left( \frac{RMSE(i)}{\mu(i)} \right)^2}$$

Where, $\frac{h}{l}$ is the ratio of pixel sizes of input PAN and MS images, $\mu(i)$ is the mean of the $i^{th}$ band while, N is the number of bands. The desired value of ERGAS is as close to zero as possible.

### 4.2.2. Spatial quality assessment

To assess the spatial quality of the resultant pansharpened image, several quantitative metrics are suggested in [38]. Zhou at.el [38], suggested spatial correlation coefficient (SCC) to measure the spatial quality of the image using laplacian filter. Spatial quality of the final pansharpened image F is assessed against input PAN image. The laplacian filter is applied to the both F and P images. SCC is calculated by obtaining the correlation between images. In [39], another method to assess the spatial quality of the pansharpened image is suggested. In that, it is suggested that the good pansharpened image retains all the edge information of the PAN image. Several edge detection techniques are applied to detect the edges in the pansharpened image, and then compared with the edges of the PAN image. SCC value close to 1 indicates high spatial quality of the pansharpened image.

## 5. RESULTS AND ANALYSIS

In our experiments, geometrically registered three data sets of input PAN and MS images are considered. Performance of the several state-of-art pansharpening algorithms is observed on these three datasets. In all three datasets, PAN images are having pixel size of 512 x 512 while MS images having size of 128 x 128. MS images are resized to concerned PAN image by using interpolation technique before applying pansharpening algorithms. Pansharpening algorithms discussed in Section 3, Brovey transform, IHS, adaptive-IHS, PCA and Discrete wavelet transform (DWT) are implemented. To compare the performance of each algorithm, subjective and objective assessment techniques are applied. Visual inspection technique is applied to observe final pansharpened image for subjective assessment. But, it is difficult to match the colors of pansharpened image to the original MS image. In objective assessment, Correlation coefficient (CC), RMSE, RASE, universal quality index Q, and ERGAS parameters are calculated to estimate spectral quality while spatial-CC (SCC) is computed to approximate spatial quality. In our experiments, geometrically registered three data sets images are considered. Several state-of-art pansharpening algorithms are implemented and results are observed. Image-1(a) and (b) shows worldview satellite urban area and seaside MS images respectively while Image-1(c) shows Quickbird satellite MS image. Image 2(a-c) shows the corresponding PAN images.

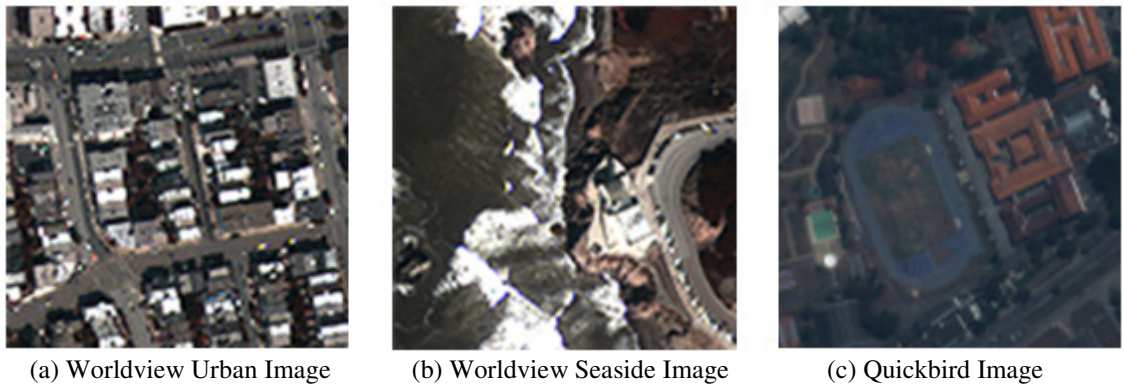

    (a) Worldview Urban Image        (b) Worldview Seaside Image        (c) Quickbird Image

Figure 1. Multispectral Images

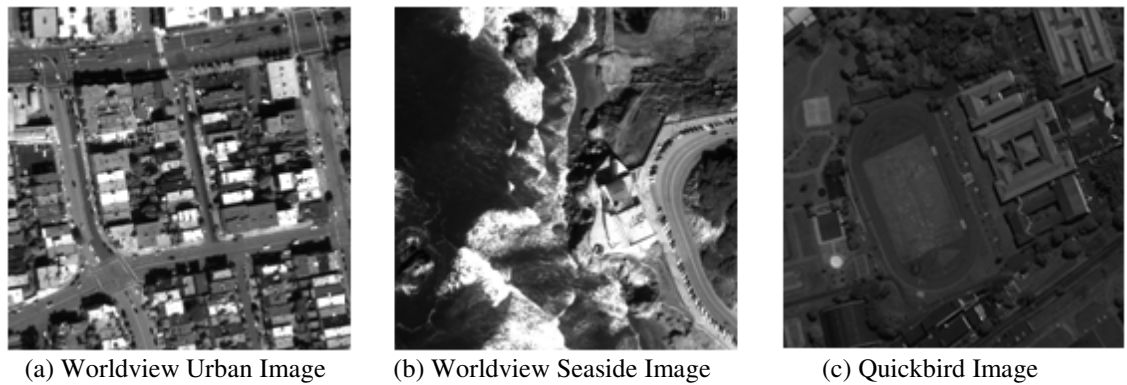

    (a) Worldview Urban Image        (b) Worldview Seaside Image        (c) Quickbird Image

Figure 2. Panchromatic Images

In all three datasets, PAN images are having pixel size of 512 x 512 while MS images having size of 128 x 128. In pre-processing MS images are resized to that of PAN images by using interpolation technique. Brovey transform, IHS, adaptive-IHS, PCA and Discrete wavelet transform (DWT) are implemented. Quantitative assessment parameters for various pansharpened algorithms for all three datasets are calculated and shown in the Tables 1. In the calculation of ERGAS parameter of each dataset, PAN and MS image pixel size ratio is considered as ¼.

Table 1. Quantitative assessment results.

| Image-1:Worldview urban area image | | | | |
|---|---|---|---|---|
| | **Brovey** | **IHS** | **Adaptive-IHS** | **PCA** | **DWT** |
| CC | 0.8909 | 0.8922 | 0.8941 | 0.8917 | **0.9306** |
| ERGAS | 4.1140 | 7.1312 | 7.0991 | 8.3854 | **6.0464** |
| Quality | 0.8904 | 0.8922 | 0.8940 | 0.7789 | **0.9282** |
| RASE | 28.5199 | 28.5126 | 28.4608 | 33.5569 | **24.1675** |
| RMSE | 26.4290 | 26.4381 | 26.3901 | 31.1155 | **22.4092** |
| SCC | 0.9907 | **0.9986** | 0.9815 | 0.9862 | 0.9095 |
| Image-2:Worldview seaside image | | | | |
| | **Brovey** | **IHS** | **Adaptive-IHS** | **PCA** | **DWT** |
| CC | 0.8286 | 0.8288 | 0.8759 | 0.8283 | **0.9393** |
| ERGAS | 8.1074 | 7.6741 | 6.4912 | 8.0171 | **4.6738** |
| Quality | 0.8277 | 0.8288 | 0.8758 | 0.7213 | **0.9387** |
| RASE | 31.0202 | 30.6398 | 26.0137 | 32.0135 | **18.6590** |
| RMSE | 27.2404 | 26.9064 | 22.8440 | 28.1128 | **16.3854** |
| SCC | 0.9963 | **0.9988** | 0.9170 | 0.9877 | 0.7236 |
| Image-3:Qickbird image | | | | |
| | **Brovey** | **IHS** | **Adaptive-IHS** | **PCA** | **DWT** |
| CC | 0.7335 | 0.7605 | 0.8908 | 0.8039 | **0.9522** |
| ERGAS | 5.0523 | 5.1351 | 3.4613 | 5.0980 | **2.3865** |
| Quality | 0.7237 | 0.7600 | 0.8902 | 0.7317 | **0.9512** |
| RASE | 22.6808 | 20.3282 | 13.6959 | 19.5795 | **9.4678** |
| RMSE | 13.9451 | 12.4986 | 8.4208 | 12.0383 | **5.8212** |
| SCC | 0.9435 | **0.9761** | 0.7287 | 0.8942 | 0.6996 |

The best values obtained for each parameter are highlighted in the tables. It is observed that, multi-resolution approaches (DWT) are preserving better spectral information while component substitution approach (IHS) improves spatial quality of input MS image. Resultant fused images are shown in figure 2.

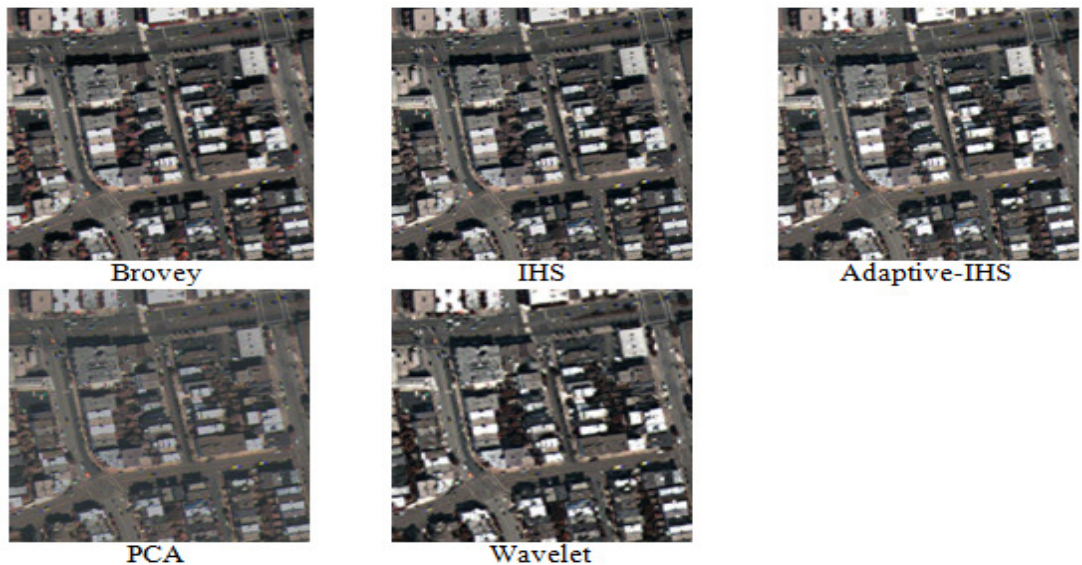

Figure-2(a). Worldview urban area Fused images

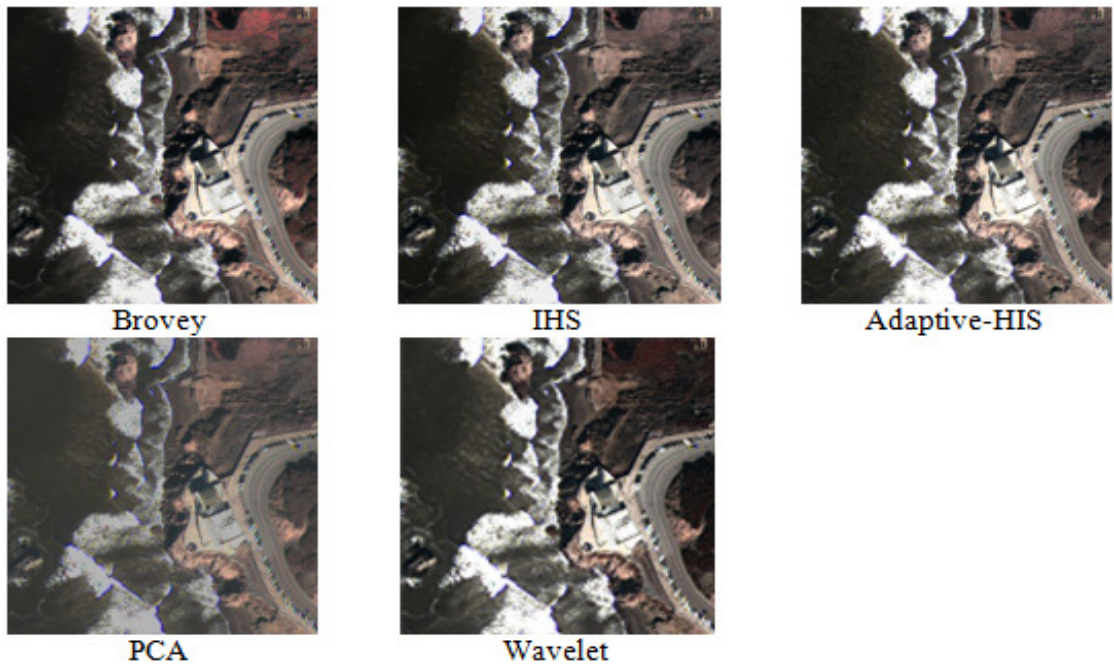

Figure-2(b). Worldview seaside Fused images

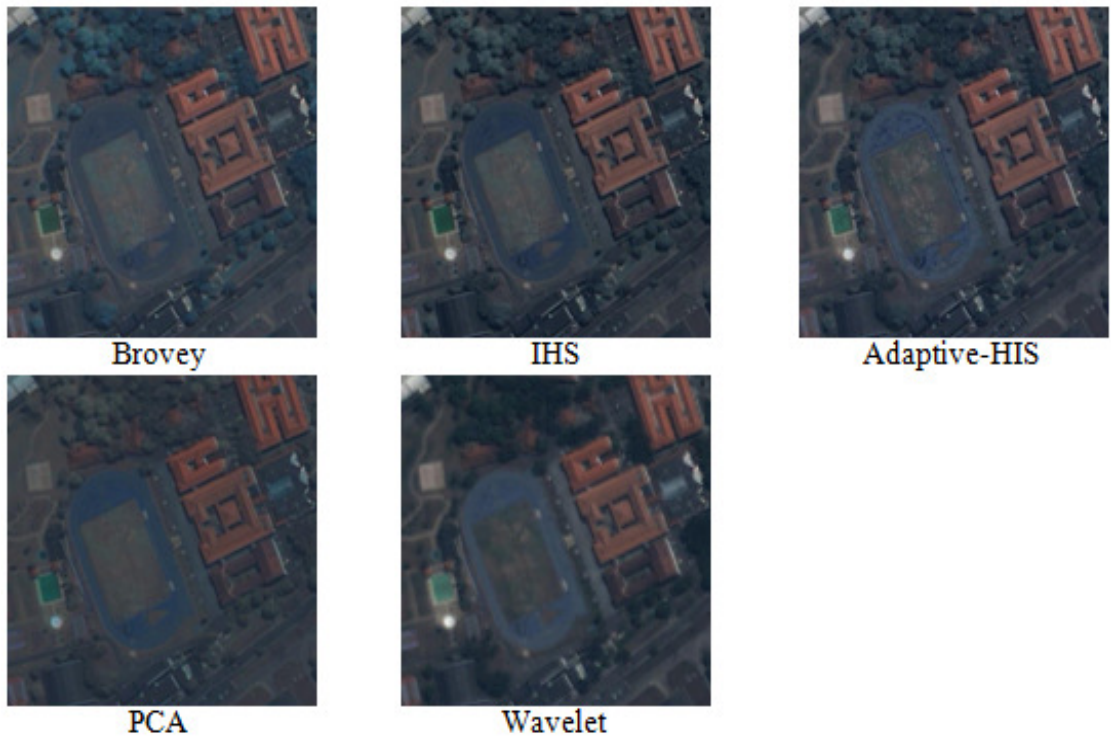

Figure-2(c). Quickbird Fused images

# 6. CONCLUSIONS

In this paper, various pansharpening techniques with their classification are discussed. They are classified based on the approach they have been using. IHS is classical CS based pansharpening technique, and its major drawback is the spectral distortion which it introduces during pansharpening process. The reason for the spectral distortion appears to may be the large radiometric difference between I and PAN bands. It could be overcome by computing the high spatial resolution image I which will ultimately reduce the difference between I and PAN bands. In PCA, first principal component of image is replaced with histogram normalized panchromatic (PAN) image. The first principal component has a largest variance and therefore, it contains most of the information. The remaining principal components possess band specific information and they are kept unaltered. One of the possible research issues is optimal replacement of principal component/s with PAN image. In spectral contribution-based approach, Brovey transform (BT) works well when images contain three bands. It preserves spectral information in the resultant pansharpened image. In statistical methods, it is desirable to estimate the accurate model for the relationship between pansharpened image, and input MS and PAN images. It is observed that multi-resolution based pansharpening approach generate better results.

## ACKNOWLEDGEMENT

Author is thankful to Prof.Tanish zaveri, Associate professor, Nirma University for his support to collect datasets.

## REFERENCES

[1] Satellite imaging corporation web page: http://www.satimagingcorp.com/gallery

[2] J.G.Liu, "Smoothing filter-based intensity modulation: A spectral preserve image fusion technique for improving spatial details", International journal of remote sensing, Vol 21, issue 18, pp.3461-3472, 2000

[3] The Online Resource for Research in Image Fusion website http://www.imagefusion.org

[4] Pascal Sirguey, Renaud Mathieu, Yves Arnaud, Muhammad M. Khan and Jocelyn Chanussot, "Improving MODIS spatial resolution for snow mapping using wavelet fusion and ARSIS concept", IEEE Geoscience and Remote Sensing Letters, Vol. 5, No. 1, pp. 78-82, 2008.

[5] Francesca Bovolo, Lorenzo Bruzzone, Luca Capobianco, Andrea Garzelli, Silvia Marchesi and Filippo Nencini, "Analysis of the effects of pansharpening in change detection on VHR images", IEEE Geoscience and Remote Sensing Letters, Vol. 7, No. 1, pp. 53-57, 2010.

[6] M. A. Ning, Ze-Ming ZHOU, Peng ZHANG and Li-Min LUO, "A new variational model for panchromatic and multispectral image fusion", Acta Automatica Sinica, Vol. 39, No. 2, pp. 179-187, 2013.

[7] Bruno Aiazzi, Stefano Baronti and Massimo Selva, "Improving component substitution pansharpening through multivariate regression of MS+ Pan data", IEEE Transactions on Geoscience and Remote Sensing, Vol. 45, No. 10, pp. 3230-3239, 2007.

[8] Qing Guo and Shutian Liu, "Performance analysis of multi-spectral and panchromatic image fusion techniques based on two wavelet discrete approaches", Optik-International Journal for Light and Electron Optics, Vol. 122, No. 9, pp. 811-819, 2011.

[9] Wenfeng Zhan, Yunhao Chen, Jinfei Wang, Ji Zhou, Jinling Quan, Wenyu Liu and Jing Li, "Downscaling land surface temperatures with multi-spectral and multi-resolution images", International Journal of Applied Earth Observation and Geoinformation, Vol. 18, pp. 23-36, 2012.

[10] B.Zitovo and J. Flusser, "Image registration methods: A survey", Image vision computing, vol. 21, no. 11, pp. 977-1000, 2003.

[11] RC Gonzalez, RE Woods, Digital image processing, 3rd edn. (Prentice Hall, 2008).

[12] C Pohl, JLV Genderen, "Multi-sensor image fusion in remote sensing: Concepts, methods, and applications", International journal of remote sensing, 19(5), 823–854, 1998.

[13] RA Schowengerdt, "Remote Sensing: Models and Methods for Image Processing", 3rd edn, (Orlando, FL: Academic, 1997.

[14] T Ranchln, L Wald, Fusion of high spatial and spectral resolution images: The ARSIS concept and its implementation. hotogramm Eng Remote Sens. 66, 49–61 (2000)

[15] VK Sheftigara, A generalized component substitution technique for spatial enhancement of multispectral lmages using a higher resolution data set. Photogramm Eng Remote Sens. 58(5), 561–567 (1992)

[16] W Dou, Y Chen, X Li, DZ Sui, A general framework for component substitution image fusion: An implementation using the fast image fusion method. Computers & Geosciences 33, 219–228 (2007). doi:10.1016/j. cageo.2006.06.008

[17] TM Tu, SC Su, HC Shyu, PS Huang, A new look at IHS-like image fusion methods. Inf Fusion 2(3), 177–186 (2001).

[18] Sheida R, Melissa S., Daria M, Michael M., Todd W., "An adaptive IHS Pan-sharpening Method", IEEE Geoscience and remote sensing letter 2010.

[19] VP Shah, NH Younan, RL King, An efficient pan-sharpening method via a combined adaptive PCA approach and contourlets. IEEE Trans Geoscince Remote Sens. 46(5), 1323–1335 (2008)

[20] Lindsay I Smith. A tutorial on principal component analysis. Technical report, http://www.cs.otago.ac.nz/cosc453/studenttutorials/principal components.pdf.

[21] S.Zebhi, M.R.Agha, M.T.Sadeghi, "Image fusion using PCA in CS domain", an international journal of signal & image processing (SIPIJ), vol.3, No.4, August 2012.

[22] A.Medina Javier Marcello and F.Eugenio, "Evaluation of spatial and spectral effectiveness of pixel-level fusion techniques" IEEE Geoscience and Remote Sensing Letters, 2012.

[23] Maryam Dehghani, "Wavelet-based image fusion using a-trous algorithm",Technical report, Map India Conference Poster Session.

[24] V Vijayaraj, CG O'Hara, NH Younan, "Quality analysis of pansharpened images", Proc IEEE Inernational Geoscience Remote Sens Symp IGARSS'04. 1, 20–24 (2004)

[25] JC Price, "Combining multispectral data of differing spatial resolution", IEEE Trans Geosc Remote Sens. 37(3), 1199–1203 (May 1999).

[26] J Park, M Kang, "Spatially adaptive multi-resolution multispectral image fusion", International Journal of Remote Sensing 25(23), 5491–5508 (2004).

[27] D Fasbender, J Radoux, P Bogaert, "Bayesian data fusion for adaptable image pansharpening", IEEE Transactions On Geoscience And Remote Sensing 46, 1847–1857 (2008)

[28] Frosti palsson, Johnnes R. Magnus O., "A New Pansharpening Algorithm Based on Total Variation", IEEE Geoscience and Remote sensing letters, 2013.

[29] SG Mallat, "A theory for multi-resolution signal decomposition: The wavelet representation", IEEE transactions on pattern analysis and machine Intelligence 11(7), 674–693 (1989).

[30] AL da Cunha, J Zhou, MN Do, "The non-sub sampled contourlet transform: Theory, design, and applications", IEEE Trans. Image Process. 15(10), 3089–3101 (2006)

[31] PJ Burt, EH Adelson, "The Laplacian pyramid as a compact image code", IEEE Transactions on Communications. COM-3l(4), 532–540 (1983)

[32] M González-Audícana, X Otazu, "Comparison between Mallat's and the a'trous discrete wavelet transform based algorithms for the fusion of multispectral and panchromatic images", International Journal on Remote Sens. 26(3), 595–614 (2005).

[33] Minh Do, and Martin Vetterli, "The contourlet transform: An efficient Directional Multi-resolution Image Representation", IEEE Transaction on Image processing, 14(12), 2091-2106, 2005

[34] H.Yin, Shutao Li, Leyuan Fang, "Simultaneous image fusion and super-resolution using sparse representation", Information Fusion, Elsevier journal, 2012.

[35] Y.Zhang, "Theory of compressive sensing via 〖l〗_1- minimization: A non-RIP analysis and extensions", Department of computational and applied mathematics, RICE University, technical report.

[36] L Wald, T Ranchin, M Mangolini, "Fusion of satellite images of different spatial resolutions: Assessing the quality of resulting images", Photogramm Eng Remote Sens. 63, 691–699 (1997)

[37] Wang, Z., and A.C.Bovik, "A universal image quality index", IEEE Signal Processing Letters, 9(3):81-84 (2002).

[38] J.Zhou, DL Civco, JA Silander, "A wavelet transform method to merge Landsat TM and SPOT panchromatic data", Int J remote Sens. 19(4), 743–757 (1998).

[39] PS Pradhan, RL King, NH Younan, DW Holcomb, "Estimation of the number of decomposition levels for a wavelet-based multi-resolution multisensory image fusion", IEEE Trans Geoscience Remote Sens. 44(12), 3674–3686 (2006).